\documentclass[10pt,twocolumn,letterpaper]{article}
\usepackage{wacv}
\usepackage{times}
\usepackage{epsfig}
\usepackage{graphicx}
\usepackage{amsmath}
\usepackage{amssymb}
\usepackage[accsupp]{axessibility}

\def\wacvPaperID{1744} 

\wacvfinalcopy 

\ifwacvfinal
\fi

\ifwacvfinal
\usepackage[breaklinks=true,bookmarks=false]{hyperref}
\else
\usepackage[pagebackref=true,breaklinks=true,colorlinks,bookmarks=false]{hyperref}
\fi

\ifwacvfinal\fi

\begin{document}

\title{Interacting Hand-Object Pose Estimation via Dense Mutual Attention}

\author{Rong Wang \quad\quad\quad Wei Mao \quad\quad\quad Hongdong Li\\
The Australian National University \\
{\tt\small \{rong.wang, wei.mao, hongdong.li\}@anu.edu.au}
}

\maketitle

\thispagestyle{empty}
\begin{abstract}
3D hand-object pose estimation is the key to the success of many computer vision applications. The main focus of this task is to effectively model the interaction between the hand and an object. To this end, existing works either rely on interaction constraints in a computationally-expensive iterative optimization, or consider only a sparse correlation between sampled hand and object keypoints. In contrast, we propose a novel dense mutual attention mechanism that is able to model fine-grained dependencies between the hand and the object. Specifically, we first construct the hand and object graphs according to their mesh structures.  For each hand node, we aggregate features from every object node by the learned attention and vice versa for each object node. Thanks to such dense mutual attention, our method is able to produce physically plausible poses with high quality and real-time inference speed. Extensive quantitative and qualitative experiments on large benchmark datasets show that our method outperforms state-of-the-art methods. The code is available at \url{https://github.com/rongakowang/DenseMutualAttention.git}.

\end{abstract}

\section{Introduction}

Accurate and efficient pose estimation for the scene of a hand interacting with an object from a single monocular view is desired in many applications, \emph{e.g.} extended reality (XR)~\cite{starke2019neural} and human-computer iteration (HCI)~\cite{koppula2013anticipating}. Despite that great efforts have been contributed to developing effective 3D hand pose estimation algorithms \cite{iqbal2018hand, kulon2020weakly, tang2021towards, zimmermann2021contrastive, yang2022dynamic}, joint hand-object pose estimation remains especially challenging due to the severe mutual occlusion and diverse ways of hand-object manipulation.
Methods failing to tackle the aforementioned challenges tend to produce physically implausible configurations, such as interpenetration and out-of-contact. To avoid generating undesired poses, an in-depth understanding of the correlation between the hand and the interacting object is therefore required.

\begin{figure}[htp!]
\centering
  {\includegraphics[width=0.45\textwidth]{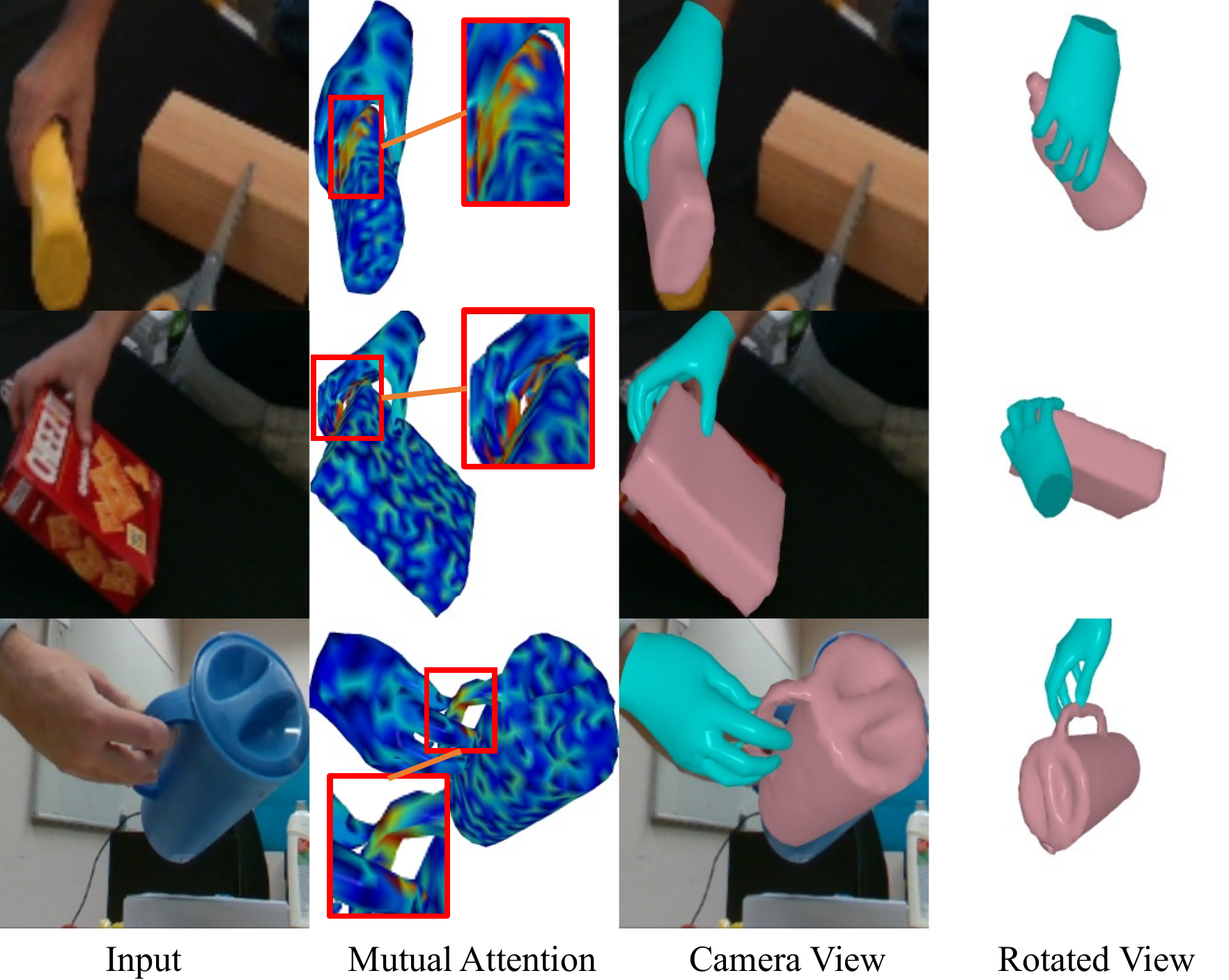}}
    \caption{\textbf{Effects of the mutual attention.} Our method recovers accurate hand-object pose via dense mutual attention between all hand and object vertices. 
    We visualize in the second column the learned average mutual attention for contacting vertices, where red regions have higher attention values and blue regions have lower values. We observe the proposed mutual attention can effectively model interaction around contacting areas. In addition, it helps to select secondary keypoints (yellow regions with medium attention values) that facilitate hand-object pose refinement.}\label{fig:ablation}
\end{figure}

Research works on 3D hand-object pose estimation can be categorized as optimization-based and learning-based. While the former methods~\cite{yang2021cpf, hasson2021towards, hampali2020honnotate} generalize to diverse object classes, the optimization process requires multiple iterations to converge, which is not applicable for real-time applications like XR. In contrast, learning-based methods~\cite{li2021artiboost, hasson2019learning, hasson2020leveraging, doosti2020hope, hampali2022keypoint} can achieve real-time inference. Motivated by the optimization-based methods, soft contact losses are introduced ~\cite{hasson2019learning, hasson2020leveraging} to implicitly guide the network to pursuit plausible hand-object interaction. For a more effective modeling, other works focus on explicitly learning the hand-object correlation~\cite{doosti2020hope, choi2017robust} in the design of the network.
Recently, several attention-based works~\cite{tse2022collaborative,hampali2022keypoint} are proposed considering its efficacy in modelling complex correlation. In~\cite{tse2022collaborative} a self-attention mechanism is used to capture feature dependencies for either the hand or the object and the interaction between them is modeled by the exchange of global features. Most close to our work is~\cite{hampali2022keypoint} where a cross-attention is used to model the correlation between the hand and the object. However, all above methods only model a \emph{sparse} interaction between a pre-defined set of keypoints or features from the hand and the object, regardless of the fact that hand-object interaction actually occurs on physical regions of the surfaces.

In this work, we instead propose to model fine-grained hand-object interaction via a \emph{dense mutual attention} mechanism. Specifically, we first estimate rough hand and object meshes separately from a single monocular image. Next, we construct the hand and object graphs based on their mesh structures, then spatially sample node features according to the rough mesh positions. Unlike \cite{tse2022collaborative} which transfers inter-graph dependencies via global features only, we allow direct node-to-node feature aggregation via mutual attention. Taking a node from the hand graph as an example, we calculate the object-to-hand attention for all object nodes, and then fuse the hand node feature with attention-weighted object node features to explicitly model the fine-grained interaction correlation. A similar calculation is performed to refine object node features given hand-to-object attention. Finally, we refine the hand and object poses through graph convolutional blocks equipped with the proposed mutual attention layer. We show that our method does not require iterative optimization as in \cite{yang2021cpf, hasson2021towards}, and the dense vertex-level mutual attention can model the hand-object interaction more effectively than sparse keypoints based methods \cite{hampali2022keypoint, doosti2020hope}. In summary, our contributions are as follows.
\begin{itemize}
    \item We propose a novel dense mutual attention mechanism that effectively models hand-object interaction by aggregating and transferring node features between the hand and object graphs.
    
    \item We design a novel hand-object pose estimation pipeline facilitating the proposed mutual attention. Extensive experiments show superior results compared to state-of-the-art methods on large benchmark datasets.
\end{itemize}


\section{Related Works}

\def \etal {\emph{et al. }}
\makeatletter
  \newcommand\incircbin
  {%
    \mathpalette\@incircbin
  }
  \newcommand\@incircbin[2]
  {%
    \mathbin%
    {%
      \ooalign{\hidewidth$#1#2$\hidewidth\crcr$#1\bigcirc$}%
    }%
  }
  \newcommand{\ostar}{\incircbin{*}}
  \newcommand{\od}{\incircbin{\cdot}}
  \newcommand{\op}{\incircbin{+}}
  \newcommand{\otime}{\incircbin{\times}}
  \makeatother

In this section, we review related works on hand-object pose estimation. Since our work relies on graph convolutional networks and the attention mechanism, we also review their utilization in related tasks.

\subsection{Hand-Object Pose Estimation}
Most previous works tackle 3D hand pose estimation \cite{iqbal2018hand, kulon2020weakly, tang2021towards, zimmermann2021contrastive, yang2022dynamic} and object pose estimation \cite{li2018deepim, peng2019pvnet, wang2021gdr, yin2021graph} separately. Recently joint hand-object pose estimation has received more focus \cite{hasson2019learning, li2021artiboost, liu2021semi, hasson2020leveraging, doosti2020hope, hasson2021towards, hampali2022keypoint} due to the strong correlation when hands interact with objects. For learning-based methods, Hasson~\etal\cite{hasson2019learning} propose attraction and repulsion losses to penalize physically implausible reconstructions. Shaowei~\etal\cite{liu2021semi} adopt a semi-supervised learning framework with contextual reasoning of hand and object representations. Hasson~\etal\cite{hasson2020leveraging} extend to video inputs by leveraging photometric and temporal consistency on sparsely annotated data. To tackle the lack of 3D ground truth, Kailin \etal\cite{li2021artiboost} introduce an online synthesis and exploration module to generate synthetic hand-object poses from a predefined set of plausible grasps during training. In contrast to the above works, optimization-based methods \cite{hasson2021towards, yang2021cpf, hampali2020honnotate} formulate the task by firstly estimating initial hand and object poses in isolation, then jointly refining them with contact constraints. However, these methods are time-consuming as the optimization process generally requires multiple iterations to converge, thus limiting their applications in real-time XR systems. In consequence, we adopt the learning-based framework and continue to introduce related works in this category in the following section.

\begin{figure*}[htp!]
\centering
  {\includegraphics[width=0.95\textwidth]{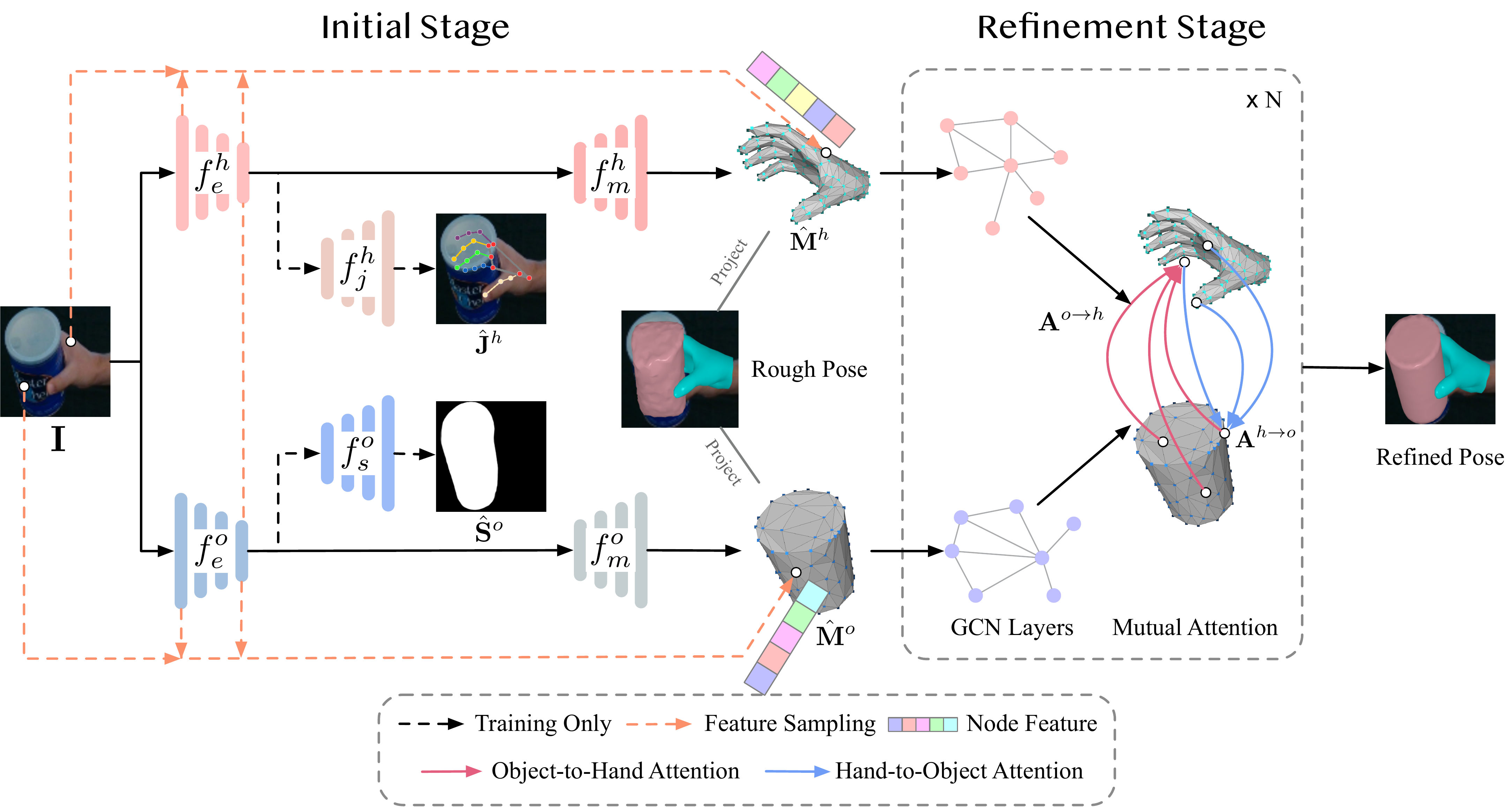}}
    \caption{\textbf{Overview of our method.} Our model consists of two stages. At the initial stage (left), we use two separate branches to estimate rough meshes of hand and object ($\hat{\mathbf{M}}^h$ and $\hat{\mathbf{M}}^o$), respectively in the mesh estimator $f_m^h(\cdot)$ and $f_m^o(\cdot)$. Each estimator takes image features from the encoder $f_e^h(\cdot)$ and $f_e^o(\cdot)$ as the input. To supervise the feature extraction, we include two additional estimators ($f_j^h(\cdot)$ and $f_s^o(\cdot)$) to estimate 3D hand joints ($\hat{\mathbf{J}}^h$) and object silhouette ($\hat{\mathbf{S}}^o$) during training. At the refinement stage (right), we first construct hand and object graphs according to the mesh structures. The initial feature of each node in the graphs is sampled from the input image $\mathbf{I}$ as well as feature maps of the image encoders according to the pixel location projected from the meshes. Finally, we leverage a stack of GCN layers followed by the proposed mutual attention layer to generate the refined hand and object poses.}
    \label{fig:main}
\end{figure*}

\subsection{GCNs-based Methods}
Graph Convolutional Networks (GCNs) have been wildly applied in 3D hand pose estimation \cite{ge20193d, tang2021towards, khaleghi2021multi, Chen_2021_WACV} since hand meshes and kinematic trees naturally form a graph. Several works have extended GCNs to hand-object pose estimation and achieved promising results. Bardia~\etal\cite{doosti2020hope} build an adaptive Graph-UNet (HOPE-Net) combining hand joints and object bounding box corners with learnable adjacent matrices. Lin~\etal\cite{huang2020hot} encode initial 2D poses with GCNs similar to HOPE-Net as priors for the following 3D reconstruction in a non-autoregressive Transformer. However, the aforementioned methods only construct sparse graphs from hand-object interaction scenes and do not estimate hand shapes, thus lacking expressiveness. Tze~\etal \cite{tse2022collaborative} propose a collaborative method to iteratively refine results from dense hand and object graphs. However, the iterative refinement is computationally expensive, and the model-free approach in object representation often does not recover accurate object shapes.

\subsection{Attention-based Methods}
Attention mechanism~\cite{vaswani2017attention} has shown remarkable success in human body~\cite{chu2017multi, kocabas2021pare} and hand pose~\cite{park2022handoccnet} estimation as it can effectively model long-range correlation and aggregate component features. Hampali \etal \cite{hampali2022keypoint} propose to learn attention between a sparse set of sampled hand and object keypoints. In \cite{tse2022collaborative}, an attention-guided GCN is proposed to effectively aggregate vertex features within either hand or object graphs. The interaction between the hand and the object is explored via the exchange of global features during the iterative process. In contrast, we propose to exploit mutual attention between every hand and object vertex that better learns the interaction dependencies.

\section{Methods}
In this section, we introduce the training pipeline as shown in Figure \ref{fig:main}. Our model consists of two stages. At the initial stage, we first separately estimate the rough hand (Section \ref{hand_pose}) and object pose (Section \ref{object_pose}) given an input RGB image $\mathbf{I} \in \mathbb{R}^{H \times W \times 3}$. Combining rough poses from both branches, we then jointly refine them at the refinement stage using a graph convolution network equipped with the proposed mutual attention layer (Section \ref{gcn}) to explicitly model the hand-object interaction. The final outputs of the refinement stage are the 3D vertex coordinates of the hand mesh defined in the MANO~\cite{romero2022embodied} model and the 6D object pose in $SE(3)$ that transforms the object CAD model into the camera frame. We train the proposed model end-to-end with a multi-task training objective (Section \ref{training_objective}). For the consistency of notations, we use the superscript $h$ and $o$ to indicate the hand and  object branch respectively.

\subsection{Hand Pose Estimation}\label{hand_pose}
Following~\cite{moon2020i2l}, we propose to represent a hand mesh via \emph{lixels}.
Specifically, we define the position of a 3D vertex $\mathbf{x} = [u, v, z]^T \in \mathbb{R}^3$ as its projected pixel coordinates ($u, v$) and depth ($z$). We then quantize the pixel coordinates and depth and into 3 independent 1D heatmap vectors ($\mathbf{l}_u,\mathbf{l}_v,\mathbf{l}_z$), where $\mathbf{l}_u,\mathbf{l}_v, \mathbf{l}_z\in \mathbb{R}^L$.
After scaling and normalization via the softmax operation, each entry (known as a lixel) of the heatmap vectors represents the probability of the pixel location or the depth for the vertex. Given such lixels, the vertex position can be computed with the soft-argmax \cite{chapelle2010gradient} operation as:
\begin{align}
    u &= \frac{W}{L}\cdot\text{soft-argmax}(\mathbf{l}_u)\;, \label{eq:lx1} \\
    v &= \frac{H}{L}\cdot\text{soft-argmax}(\mathbf{l}_v)\;, \label{eq:lx2} \\ 
    z &= \frac{2D}{L}\cdot\text{soft-argmax}(\mathbf{l}_z) + r_z - D \label{eq:lx3} \;,
\end{align}
 where $W$ and $H$ are the width and height of the image. $L$ is the quantization level. $D$ is the depth radius\footnote{We therefore quantize the depth ranging in [$r_z - D$, $r_z + D$].} relative to the wrist joint estimated from the training data, and $r_z$ is the wrist joint depth\footnote{Relative depth in the object mesh also refers to the wrist joint.}, which is assumed to be known~\cite{li2021artiboost, tang2021towards} to resolve the scale ambiguity in the single view input. 
 
 Given the camera intrinsic $\mathbf{K}$, pixel coordinates, and depth, we can easily recover the 3D vertex's Euclidean coordinates in the camera space. As shown in \cite{moon2020i2l}, such representation is more robust and effective than directly regressing 3D coordinates, and is more memory-efficient than 3D voxel representation as it decouples the three components. Unless otherwise specified, throughout the rest sections, our model will produce the 3 vectors ($\mathbf{l}_u,\mathbf{l}_v,\mathbf{l}_z$) when estimating mesh vertices and hand joints. Those vectors will then be converted to the vertex position $(u, v, z)$ using the equations~\ref{eq:lx1},\ref{eq:lx2} and \ref{eq:lx3}.
 
 Recall that, at the initial stage, we use two separate branches to estimate rough hand and object meshes. In particular, given the input image $\mathbf{I}$, the hand pose estimation branch first extracts image features using an image feature encoder $f^h_{e}(\cdot)$:
 \begin{equation}
     \{\mathbf{F}^h_{(i)}\} = f^h_{e}(\mathbf{I})\;,
 \end{equation}
 where $f^h_{e}(\cdot)$ is implemented as a ResNet-50 \cite{he2016deep} encoder pre-trained on the ImageNet \cite{russakovsky2015imagenet} and $\{\mathbf{F}^h_{(i)}\}$ denotes the collection of feature maps extracted from the $i$-th layer of the encoder. In particular, we denote the image feature map from the final layer as $\mathbf{F}^h$ for succinct notions.

 To guide the feature extraction, we additionally feed the estimated image feature from the final layer in a hand joint estimator $f_j^{h}(\cdot)$:
 \begin{equation}
     \hat{\mathbf{J}}^h = f_j^h(\mathbf{F}^h)\;,
 \end{equation}
 where $\hat{\mathbf{J}}^h \in \mathbb{R}^{21\times 3}$ are the estimated positions of 21 hand joints. Note that, the joint estimator is only used for the purpose of feature extraction supervision in training. During testing, the entire joint estimator is removed. 
 
 Finally, given the final image feature $\mathbf{F}^h$, we obtain a rough hand mesh $\hat{\mathbf{M}}^h \in \mathbb{R}^{778 \times 3}$ from the hand mesh estimator $f_m^h(\cdot)$:
 \begin{equation}
     \hat{\mathbf{M}}^h = f_m^h(\mathbf{F}^h)\;.
 \end{equation}

\subsection{Object Pose Estimation}
\label{object_pose}
Similar to the hand pose estimation branch, we first extract image features through an image encoder $f_e^o(\cdot)$ which has the same architecture as $f_e^h(\cdot)$ but does not share weights:
\begin{equation}
     \{\mathbf{F}_{(i)}^o\} = f_e^o(\mathbf{I})\;.
\end{equation}
We also use $\mathbf{F}^o$ to denote the feature map extracted from the final layer of $f_e^o$($\cdot$). 

Since there are no unanimous keypoints defined for all classes of objects, we alternatively use the object silhouette to supervise the feature extraction. Specifically, we design the object mask estimator $f_s^o(\cdot)$ taking of the input as $\{\mathbf{F}_{(i)}^o\}$. Following the image segmentation literature~\cite{ronneberger2015u}, we include skip-connections from the image encoder to the mask estimator. Hence all image features are forwarded into the estimator to obtain the object silhouette $\hat{\mathbf{S}}^o\in\mathbb{R}^{H\times W}$ as:
\begin{equation}
    \hat{\mathbf{S}}^o = f_s^o(\{\mathbf{F}_{(i)}^o\})\;.
\end{equation}

Similarly, we construct the object mesh estimator $f_m^o(\cdot)$ symmetric to $f_m^h(\cdot)$. When estimating the object mesh, we follow the previous work~\cite{hasson2021towards} to assume that the object CAD models are given and the meshes are resampled to have 1000 vertices using ACVD~\cite{valette2008generic} for the convenience of batch training. The object mesh $\hat{\mathbf{M}}^o \in \mathbb{R}^{1000 \times 3}$ can be computed as:
\begin{equation}
    \hat{\mathbf{M}}^o = f_m^o(\mathbf{F}^{o})\;.
\end{equation}
Note that, at the initial stage, instead of directly regressing the target rough 6D object pose, we adopt a model-free approach as used in ~\cite{hasson2019learning} to estimate rough object meshes. Empirically, we find that such a strategy is more robust and it better facilitates feature sampling introduced in the following section. 

\subsection{Hand-Object Pose Refinement}
\label{gcn}
Given the rough meshes of the hand $\hat{\mathbf{M}}^h$ and the object $\hat{\mathbf{M}}^o$, we then jointly refine them by exploiting their correlations. To this end, we regard those meshes as two graphs and propose to use the graph convolutional network (GCN)~\cite{kipf2016semi} to capture the intra-graph dependencies. To further model the inter-graph interaction, we propose a novel mutual attention layer that allows fine-grained feature aggregation between two graphs.

\noindent
\textbf{Graph Construction.} 
As shown in Figure \ref{fig:main}, the hand and the object are modeled by separate graphs with vertices as nodes and their connections defined in the mesh structures as edges. Vertices belonging to different branches are disconnected and communicate via mutual attention. Motivated by~\cite{tang2021towards}, we initialize the features of each graph node from the feature extraction module at the initial stage. Taking the hand graph for example, given the pixel coordinates of $n$-th node $\mathbf{v}_n=[u_n,v_n]^T$ in the rough mesh $\hat{\mathbf{M}}^h$, we spatially sample local features from image features $\{\mathbf{F}_{(i)}^h\}$ using a bilinear interpolation operation $f_b(\cdot)$. In the meantime, we fuse final image features from both the hand and object branches to obtain a global feature containing the global information of the hand and object mesh structures. The initial node features $\mathbf{h}_n^{h}$ is computed as a concatenation of the local and global features:
\begin{equation}
    \mathbf{h}_n^{h} = f_b(\mathbf{I}(\mathbf{v}_n))  \oplus f_b(\{\mathbf{F}_{(i)}^h(\mathbf{v}_n)\}_{i\in\mathcal{X}}) \oplus f_g(\mathbf{F}^h + \mathbf{F}^o)\;,
\end{equation}
where $\mathbf{h}_n^{h}\in\mathbb{R}^{K}$, $\mathcal{X}$ is a set of layer indices from which we sample feature maps, $f_g(\cdot)$ is a global feature fusion unit, and $\oplus$ denotes the concatenation operation.

For the $m$-th node of the object graph, we compute the initial feature $\mathbf{h}_m^{o}\in\mathbb{R}^{K}$ in a similar way:
\begin{equation}
    \mathbf{h}_n^{h} = f_b(\mathbf{I}(\mathbf{v}_m))  \oplus f_b(\{\mathbf{F}_{(i)}^o(\mathbf{v}_m)\}_{i\in\mathcal{X}}) \oplus f_g(\mathbf{F}^h + \mathbf{F}^o)\;,
\end{equation}

\noindent
\textbf{Graph Convolutional Layer. } After initializing the node features, we then follow~\cite{xu2018powerful} to update the node features via graph convolutional layers. For hand nodes, the feature updating can be expressed as:
\begin{equation}\label{eq5}
    {\mathbf{h}'_n}^h = \text{MLPs}^h(\mathbf{h}_n^h + \sum_{i \in \mathcal{N}_n} \mathbf{h}_i^h)\;,
\end{equation}
where $\mathcal{N}_n$ is the indices of neighboring nodes to the $n$-th node and $\text{MLPs}^h$ denotes several sequential multi-layer perceptrons. Updating object node features follows the same in equation \ref{eq5} by changing the superscript $h$ to $o$. Intuitively, the graph convolutional layers exploit neighboring correlation from the topology of the mesh model and thus, can effectively model intra-graph dependencies.

\noindent\textbf{Mutual Attention Layer. } As shown in Figure~\ref{fig:main}, following one or several graph convolutional layers, we model hand-object interaction in the mutual attention layer. For each node from one graph, our mutual attention layer aims to aggregate features from the other graph via the attention mechanism. Specifically, for every node feature in the hand graph, we first use three 1D convolutional layers to extract the query, key, and value, and collect all queries, keys, and values as $\mathbf{Q}^{h}\in\mathbb{R}^{778\times H}$, $\mathbf{K}^{h}\in\mathbb{R}^{778\times F}$ and $\mathbf{V}^{h}\in\mathbb{R}^{778\times F}$ respectively, where each row of them is the query, key or value of a particular node. Similarly, we have the query, key and value for the object graph as $\mathbf{Q}^o\in\mathbb{R}^{1000\times F}$, $\mathbf{K}^o\in\mathbb{R}^{1000 \times F}$ and $\mathbf{V}^o\in\mathbb{R}^{1000 \times F}$ respectively. We then compute the object-to-hand attention between the queries from the hand graph and the keys from the object graph following \cite{vaswani2017attention} as:
\begin{equation}
    \mathbf{A}^{o\rightarrow h} = \text{softmax}(\frac{\mathbf{Q}^h{\mathbf{K}^o}^T}{\sqrt{F}})\;,
\end{equation}
where $\mathbf{A}^{o\rightarrow h} \in \mathbb{R}^{778\times 1000}$ is the object-to-hand attention map, with the $i$-th row denoting the expected contribution proportion of all object nodes to the $i$-th hand node. The softmax operation is performed along the second dimension. We can then aggregate object node features weighted by the object-to-hand attention as:
\begin{equation}
    \mathbf{h}^{o \rightarrow h} = \mathbf{A}^{o\rightarrow h} \mathbf{V}^{o}\;,
\end{equation}
where $\mathbf{V}^{o \rightarrow h}\in\mathbb{R}^{778\times F}$ is the aggregated features from the object graph. Similarly, we can compute the hand-to-object attention as:
\begin{equation}
    \mathbf{A}^{h\rightarrow o} = \text{softmax}(\frac{\mathbf{Q}^o{\mathbf{K}^h}^T}{\sqrt{F}})\;,
\end{equation}
where $\mathbf{A}^{h\rightarrow o}\in\mathbb{R}^{1000\times 778}$. And we can compute the hand-to-object feature as:
\begin{equation}
    \mathbf{h}^{h \rightarrow o} = \mathbf{A}^{h\rightarrow o} \mathbf{V}^{h}\;,
\end{equation}
where $\mathbf{V}^{h \rightarrow o}\in\mathbb{R}^{1000\times F}$. We finally fuse the aggregate feature with the original feature in each node as:
\begin{equation}
    \tilde{\mathbf{h}}_n^h = f_v^h({{\mathbf{h}}'_n}^h \oplus \mathbf{h}^{{o \rightarrow h}})\;, \quad \tilde{\mathbf{h}}_n^o = f_v^o({{\mathbf{h}}'_n}^o \oplus \mathbf{h}^{{h \rightarrow o}})\;.
\end{equation}
where $\tilde{\mathbf{h}}_n^h$ is the refined node feature as the output of each block, and $f_v^h(\cdot)$, $f_v^o(\cdot)$ are independent fusion units.

Intuitively, the mutual attention encodes the feature similarities between object and hand features. Since the local features are retrieved from the interpolation in the spatial domain, we expect vertices that are spatially close should be encoded with similar features due to the averaging effect in the interpolation. In this sense, the attention mechanism can effectively exploit interaction priors around contacting areas, as illustrated in Figure \ref{fig:ablation}. In addition, since we evaluate the mutual attention between every pair of hand and object vertices, this process also allows for fine-grained hand-object interactions, which as will be shown in the experiment section, performs better than methods with only attention between sparse keypoints ~\cite{hampali2022keypoint}.

\noindent
\textbf{Refined Pose.} The final output of the hand GCN is a mesh vertex offset $\Delta \mathbf{M}^h\in\mathbb{R}^{778\times 3}$, the refined hand mesh is then $\tilde{\mathbf{M}}^h = \hat{\mathbf{M}}^h + \Delta \mathbf{M}^h$. The object GCN outputs a 6D pose including the rotation and translation. In particular, inspired by \cite{yin2021graph} the object GCN extracts one object pose from every node of the object graph and the final pose ($\hat{\mathbf{R}}^o,\hat{\mathbf{T}}^o$) is the average across all poses. We empirically found this gives a better pose than only estimating one pose from the entire graph.

\subsection{Training Objectives}
\label{training_objective}
To effectively train the proposed model, we adopt a multi-tasking training objective. We first adopt an L1 loss to supervise rough and refined mesh predictions as: 
\begin{equation}
    \mathcal{L}_m = || \hat{\mathbf{M}}^h - \mathbf{M}^h ||_1 + 
    || \tilde{\mathbf{M}}^h - \mathbf{M}^h ||_1 + || \hat{\mathbf{M}}^o - \mathbf{M}^o ||_1\;,
\end{equation}
where $\mathbf{M}^h$ and $\mathbf{M}^o$ denote the ground truth mesh for the hand and the object respectively. Following \cite{tang2021towards}, we further refine the mesh quality by imposing the edge loss $\mathcal{L}_e$ and the normal loss $\mathcal{L}_n$ to penalize flying vertices and irregular surfaces as:
\begin{align}
  \mathcal{L}_e = &\sum_{i}\| |\hat{\mathbf{e}}_i^h| - |\mathbf{e}_i^h| \|_1 + 
    \| |\tilde{\mathbf{e}}_i^{h}| - |\mathbf{e}_i^h| \|_1 \nonumber \\
    &+ \sum_{j}\| |\hat{\mathbf{e}}_j^o| - |\mathbf{e}_j^o| \|_1\;,
\end{align}
\begin{align}
  \hspace*{-10cm}\mathcal{L}_n = &\sum_{i}\| \langle \hat{\mathbf{e}}_i^h, \mathbf{n}_{i}^{h} \rangle ||_1 + 
    \| \langle \tilde{\mathbf{e}}_i^h, \mathbf{n}_{i}^{h} \rangle \|_1 \nonumber \\ 
    &+ \sum_{j}\| \langle \hat{\mathbf{e}}_j^o, \mathbf{n}_{i}^{o} \rangle \|_1\;,
\end{align}
where $\hat{\mathbf{e}}_i^h$ and $\tilde{\mathbf{e}}_i^h$ denote the $i$-th mesh edge vector of the rough hand mesh and the refined hand mesh respectively. $\hat{\mathbf{e}}_j^o$ is the $j$-th mesh edge of the rough object mesh. $|\cdot|$ represents the length of the edge. $\mathbf{e}_i^h$ and $\mathbf{n}_{i}^{h}$ are the ground truth edge vector and the normal of the corresponding edge.

To supervise the refined object pose, we adopt an L2 loss on the estimated rotation quaternion and translation as:
\begin{equation}
  \mathcal{L}_o = \| \hat{\mathbf{R}}^o - \mathbf{R}^o \|_2 + \| \hat{\mathbf{T}}^o - \mathbf{T}^o\|_2\;,
\end{equation}
where $\mathbf{R}^o$ and $\mathbf{T}^o$ denote the ground truth object pose.

To supervise the hand joint estimation, we adopt a joint loss $\mathcal{L}_j$ between the ground truth joints ${\mathbf{J}}^h$ and the predicted joints from the joint estimator $\hat{\mathbf{J}}^h$, as well as the regressed joints from the predicted hand mesh, \emph{i.e.}, we use the joint regression matrix $\mathbf{G}\in \mathbb{R}^{21\times778}$ defined in the MANO \cite{romero2022embodied} model to obtain the joint locations, then calculate the joint loss as: 
\begin{equation}
\begin{aligned}
  \mathcal{L}_j &= \| \mathbf{G}\hat{\mathbf{M}}^h - \mathbf{J}^h \|_1 + 
    \| \mathbf{G}\tilde{\mathbf{M}}^h - \mathbf{J}^h ||_1 + || \hat{\mathbf{J}}^h - \mathbf{J}^h \|_1\;,
\end{aligned}
\end{equation}

Besides, we also guide the prediction of the object silhouette using a cross-entropy loss as: 
\begin{equation}
  \mathcal{L}_s = -\sum_{i=1}^{H \times W} y_i \log s_i\;,
\end{equation}
where $s_i$ is the $i$-th pixel in the predicted object silhouette $\hat{\mathbf{S}}^o$ and $y_i$ is the ground truth at the same pixel. 

Finally, inspired by \cite{tang2021towards}, we impose a finger rendering loss $\mathcal{L}_f$ to supervise the alignment of fingers in the image space. We adopt a differentiable renderer $f_r(\cdot)$ \cite{kato2018neural} to render the refined hand mesh as well as the ground truth hand mesh using the given camera intrinsic $\mathbf{K}$. We then classify the type of finger for each vertex based on the maximum blending weights defined in MANO and provide a distinct color texture for each finger. The loss can be formally written as the L1 loss between the two rendered images:
\begin{equation}
  \mathcal{L}_f = ||f_r(\tilde{\mathbf{M}}^h)  - f_r(\mathbf{M}^h) ||_1\;.
\end{equation}
The overall training loss is a weighted sum of all individual loss functions, defined as:
\begin{equation}
    \mathcal{L} = \lambda_m \mathcal{L}_m + \lambda_e \mathcal{L}_e + \lambda_n \mathcal{L}_n + \lambda_o \mathcal{L}_o + \lambda_j \mathcal{L}_j + \lambda_s \mathcal{L}_s + \lambda_f \mathcal{L}_f\;,
\end{equation}
where we empirically set $\lambda_m = \lambda_e = \lambda_n = \lambda_j = 1$, $\lambda_o = 10$, $\lambda_s = \lambda_f = 100$ so that all loss terms are roughly in the same scale.

\section{Experiment Results}

In this section, we first introduce the datasets for the training (Section \ref{training_data}) and define the evaluation metrics on each dataset (Section \ref{evaluation_metrics}). We then provide the implementation details (Section \ref{implementation_details}) for our experiments, and compare the results with state-of-the-art methods both quantitatively and qualitatively (Section \ref{results}). Finally, we perform an ablation study to investigate the effects of the mutual attention layer and demonstrate the learned interaction from the estimated attention maps (Section \ref{ablation}).

\subsection{Training Data}
\label{training_data}
\noindent
\textbf{Datasets.} We evaluate our methods on two large-scale hand-object benchmarks: HO3D v2 \cite{hampali2020honnotate} and DexYCB \cite{chao2021dexycb}, each containing 66K and 589K images of human interacting YCB \cite{calli2015ycb} objects. We train the model separately on each dataset based on the official train-test split, in particular, we use the default S0 split for the DexYCB testing set. For a fair comparison in the DexYCB dataset, we follow \cite{tse2022collaborative} to select input frames where the hand and object are both visible with an in-between distance less than $1cm$ to ensure a physical contact can be established. We crop the input images in both datasets using the provided hand-object bounding box following \cite{li2021artiboost} and resize all images into $256 \times 256$ pixels.\\

\begin{table*}[htp!]
\centering
  \caption{\textbf{Quantitative comparison on the HO3D v2 testing set.} Best results are highlighted in \textbf{bold} and unavailable results are marked with "-". Additional object metrics for \cite{li2021artiboost, hampali2022keypoint} are compared and included in the supplementary material.}
       \resizebox{0.95\textwidth}{!}{
       
  \begin{tabular}{l|cccc|cc|cc}
    \hline
                                        & \multicolumn{4}{c|}{Hand} & \multicolumn{2}{c|}{Object} &
                              \multicolumn{2}{c}{Interaction} \\
    \multicolumn{1}{l|}{Methods} & MJE ($cm$)$\downarrow$      & AUC-MJE  $\uparrow$      & MME ($cm$)$\downarrow$          & AUC-MME $\uparrow$ & MME ($cm$)$\downarrow$ & ADD-S ($cm$)$\downarrow$ & PD ($mm$)$\downarrow$  & CP (\%)$\uparrow$      \\ \hline
  Hasson \etal \cite{hasson2020leveraging} & 3.69 & 0.469 & 1.14 & 0.773 & 8.7 & 2.9 & - & -  \\ 
  Hasson \etal \cite{hasson2021towards} & 2.68 & 0.510 & 1.20 & 0.761 & 8.0 & 3.8 & 1.5  & 77.5 \\
  Keypoint Trans. \cite{hampali2022keypoint} & 2.57 & 0.532 & - & - & - & - & -  & - \\
  Artiboost \cite{li2021artiboost} & 2.53 & 0.532 & 1.09 & 0.782 & - & - & -  & - \\
  Ours & \textbf{2.38} & \textbf{0.560} & \textbf{1.06} & \textbf{0.789} & \textbf{5.7} & \textbf{2.3} & \textbf{1.3} & \textbf{85.6}   \\ \hline
  
  \end{tabular}}

  \label{table:ho3d}
 \end{table*}

\begin{figure*}[htp!]
\centering
  {\includegraphics[width=0.95\textwidth]{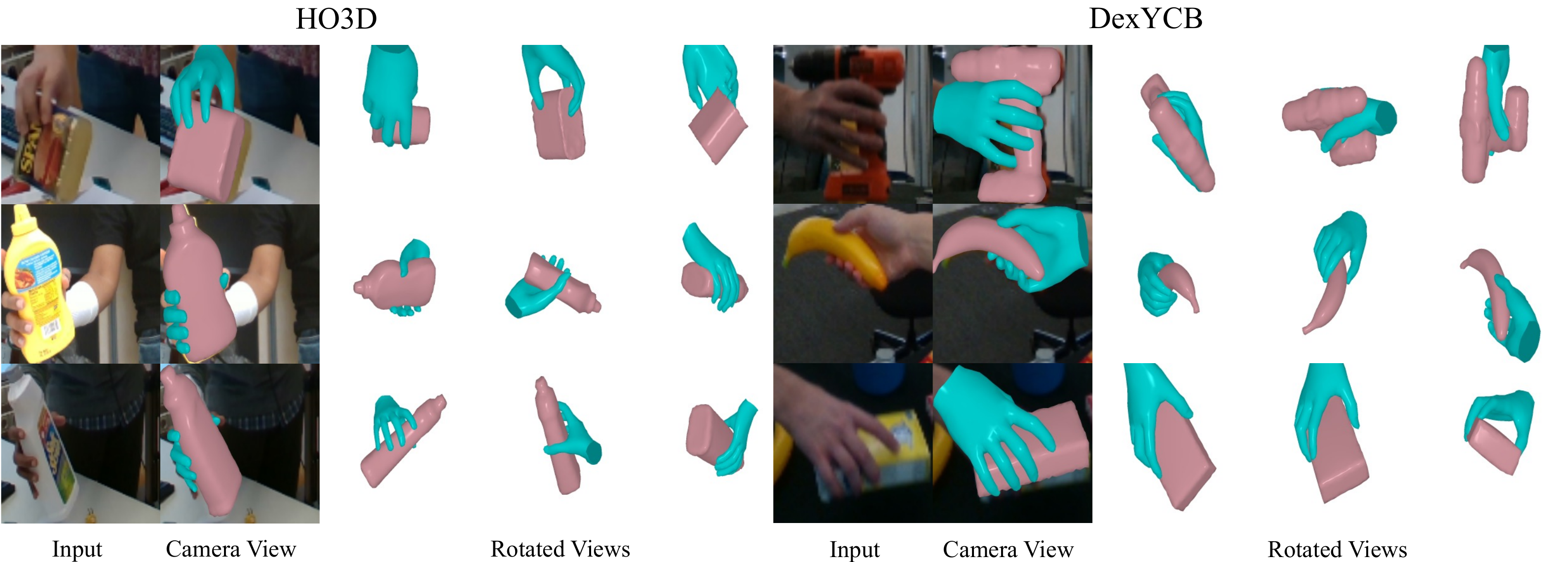}}
    \caption{\textbf{Qualitative results on the HO3D and DexYCB testing sets.} The predicted hand and object pose well align with input images in the camera view. Rotated view images show the grasping configuration is physically plausible and valid contacts can be established. }
    \label{fig:quali}
\end{figure*}

\noindent
\textbf{Data Augmentation. }Considering that the HO3D dataset is relatively small-scale, to facilitate the training we perform two types of augmentation, \emph{i.e.} view synthesis to resolve occlusion ambiguity and grasp synthesis to increase the diversity of hand-object interaction. For view synthesis, we randomly rotate the camera relative to the object center. We additionally generate 5K distinct hand manipulating YCB objects scenes using the GrabNet \cite{taheri2020grab} to perform grasp synthesis. We manually verify that the synthesized poses are not seen in the testing set and are physically plausible by empirically filtering out samples with the contact loss and penetration loss \cite{hasson2019learning} greater than the threshold $\lambda_c = 0.012$ and $\lambda_p = 0.1$ respectively. 

We use Pytorch3D \cite{ravi2020accelerating} to render the synthetic image counterparts from the augmented poses. We adopt the HTML \cite{qian2020html} model for realistic hand skin colors and textures, and superimpose the rendered hand-object images on top of randomly sampled backgrounds from the indoor-scene dataset \cite{quattoni2009recognizing}. 
To reduce the domain gap, we further perform photometric augmentation on rendered images, including random contrast and brightness transforms uniformly sampled from [0.5, 1.5]. In addition, we add random Gaussian blurs on synthetic images with $\sigma$ uniformly sampled from [0.1, 1].
The three types of inputs, \emph{i.e.} real, view-synthetic, and grasp-synthetic images are distributed in 0.45: 0.45: 0.1 in a training batch for the HO3D training set. Compared with \cite{li2021artiboost}, we introduce the data augmentation mostly with a simple view transformation, however, our model achieves better performance with less augmented grasping data as shown in Section \ref{results}.

\subsection{Evaluation Metrics}
\label{evaluation_metrics}
In order to consistently compare the results with state-of-the-art methods, we adopt the evaluation metrics on each benchmark dataset that are majorly reported by related works. We refer readers in the supplementary materials for additional metrics reported in some works \cite{li2021artiboost, hampali2022keypoint}.

\noindent
\textbf{HO3D Metrics. }For the hand pose evaluation, we follow the official evaluation metrics in the HO3D v2 CodaLab Challenge. Specifically, we report the mean joint error (MJE) \cite{zimmermann2017learning} and mean mesh error (MME) \cite{zimmermann2019freihand} as the average Euclidean distance between predicted and ground truth joints/meshes after root joint and global scale alignment. In addition, we report the AUC of the percentage of correct keypoints (PCK) curve in an interval from $0cm$ to $5cm$ with 100 equally spaced thresholds. For the object pose evaluation, we follow \cite{tang2021towards} to report the MME for the object mesh and the standard pose estimation average closest
point distance (ADD-S) \cite{xiang2017posecnn}. Finally, we report the mean penetration depth (PD) \cite{brahmbhatt2020contactpose} and the contact percentage (CP) \cite{karunratanakul2020grasping} between the hand and object meshes to evaluate the hand-object interaction.\\

\noindent
\textbf{DexYCB Metrics. }We adopt the evaluation metrics used in recent works \cite{li2021artiboost, tse2022collaborative} for the DexYCB dataset. Specifically, for the hand pose, we also report the mean joint error (MJE). For the object pose, 
 We report the mean corner error (MCE) as the distance of the  bounding box corners positions between the predicted and ground truth object meshes. Finally, we report the mean penetration depth to evaluate the hand-object collision as well.

\subsection{Implementation Details}
\label{implementation_details}
We train the network using the Adam \cite{kingma2014adam} optimizer with $\beta_1 = 0.9$ and $\beta_2 = 0.999$ on a single NVIDIA RTX 3090 GPU. We set the batch size as 24 and train the model in 25 epochs. The initial learning rate is set as $1e^{-4}$ and decayed by 0.1 after every 10 epochs. Our model achieves an inference speed of 34 FPS on an NVIDIA RTX 3090 GPU, which can be severed for future real-time applications. We refer the readers in the supplementary material for the detailed network architecture for each module.

\subsection{Results}
\label{results}
\noindent
\textbf{Comparison with State-of-the-Arts.} In Table \ref{table:ho3d}, we evaluate our model on the HO3D v2 testing set and compare the results with state-of-the-art methods \cite{hasson2020leveraging, hasson2021towards, li2021artiboost, hampali2022keypoint}. All results under hand metrics are collected from the official HO3D v2 CodaLab Challenge outcomes. From the table, we observe that our method achieves superior results across all hand, object, and interaction metrics. In particular, our method not only produces more accurate hand and object poses, but also generates physically realistic hand-object grasping in higher quality as we observe a lower penetration and a higher contact rate than \cite{hasson2021towards}. Meanwhile, our method leverages an efficient feed-forward pipeline from a single image input and does not require computationally-expensive optical flows as temporal clues \cite{hasson2020leveraging} or iterative optimization process \cite{hasson2021towards}. Furthermore, our method does not rely on sophisticated contact losses as in \cite{hasson2020leveraging, hasson2021towards}, showing the superiority of our method in modeling hand-object interaction. Compared to \cite{li2021artiboost}, our model is trained with remarkably less augmented data, yet achieves improved results without introducing much complexity. Finally, thanks to the dense mutual attention, our method improves the performance by a large margin than the sparse keypoints-based method \cite{hampali2022keypoint}.

To further justify the effectiveness of the model, we also evaluate our model on the recently released DexYCB dataset and compare the results with \cite{hasson2019learning, hasson2021towards, tse2022collaborative} in Table \ref{table:dex}. Note that while \cite{hasson2021towards} has the same setting with us, \cite{hasson2019learning, tse2022collaborative} do not assume known object CAD models, therefore tackle a more challenging task and can perform worse in estimating accurate object meshes. Hence we only compare with them in the hand metric. The results show that our method consistently outperforms baseline methods in all comparable metrics.

\begin{table}[htp!]
\centering
  \caption{\textbf{Quantitative comparison on the DexYCB testing set.} Best results are highlighted in \textbf{bold} and non-comparable results are marked with "-".}
       \resizebox{0.4\textwidth}{!}{
  \begin{tabular}{l|c|c|c}
    \hline
                                        & \multicolumn{1}{c|}{Hand} & \multicolumn{1}{c|}{Object} &
                              \multicolumn{1}{c}{Interaction}\\
    \multicolumn{1}{l|}{Methods} & MJE ($cm$)$\downarrow$    & MCE ($cm$)$\downarrow$ & PD ($mm$)$\downarrow$         \\ \hline
  Hasson \cite{hasson2019learning} & 1.76 & - & - \\
  Hasson \cite{hasson2021towards} & 1.88 & 5.25 & 0.79  \\
  Tze \etal \cite{tse2022collaborative} & 1.53 & - & - \\
  Ours & \textbf{1.27} & \textbf{3.26} & \textbf{0.67} \\ \hline
  \end{tabular}}
  \label{table:dex}
 \end{table}

\noindent
\textbf{Qualitative Results.} We show qualitative results on the HO3D and DexYCB testing sets in Figure \ref{fig:quali}. We render the estimated hand and object meshes under the camera view along with three randomly rotated views. It can be seen that our method produces accurate hand-object poses that align well with the given image input, and the estimated poses satisfy physical constraints, \emph{i.e.} a valid grasping can be observed. More results can be found in the supplementary material.
 
\subsection{Ablation Study}
\label{ablation}
To further justify the effectiveness of the proposed mutual attention mechanism, we further perform an ablation study. We first visualize the attention maps in Figure \ref{fig:ablation} (second column). For object-to-hand attention, we select hand vertices whose minimal distance to the object is less than $1cm$, and visualize the corresponding average attention between all object vertices. The hand-to-object attention is visualized in a similar way for contacting object vertices. The Figure shows that contacting areas contain higher attention values (in red) than non-contacting areas (in blue), which illustrates that the mutual attention mechanism can effectively model hand-object interaction correlation to facilitate pose refinement by exploiting contact priors.

\begin{table}[htp!]
\centering
  \caption{\textbf{Effects of the mutual attention layer}. Best results are highlighted in \textbf{bold}.}
      \resizebox{0.5\textwidth}{!}{
  \begin{tabular}{l|cc|c|c}
    \hline
    & \multicolumn{2}{c|}{Hand} & \multicolumn{1}{c|}{Object} &
                              \multicolumn{1}{c}{Interaction}\\
    \multicolumn{1}{l|}{Methods} & MJE ($cm$)$\downarrow$ & MME ($cm$)$\downarrow$    & MME ($cm$)$\downarrow$ & PD ($mm$)$\downarrow$         \\ \hline
  w/out GCN & 2.84 & 1.29 & 13.4 & 3.6 \\
  w/out attention & 2.66 & 1.20 & 7.7 & 2.9  \\
  all edge & 2.79 & 1.34  & 8.9 &  4.3 \\ 
  w/o hand-to-object  & 2.46 & 1.10 & 6.2 & 1.4 \\ 
  w/o object-to-hand & 2.50 & 1.12 & 5.8 & 1.3 \\
  mutual attention & \textbf{2.38} & \textbf{1.09} & \textbf{5.7} & \textbf{1.3}\\ \hline
  \end{tabular}}
  \label{table:ab1}
 \end{table}

We further construct variant baselines with alternative utilization of hand-object interaction priors and compare the results in Table \ref{table:ab1}. With the GCN refinement (w/o attention), the network can effectively improve hand and object pose estimation from the rough stage (w/o GCN) by a large margin thanks to the information from intra-graph dependencies.

A naive baseline method for hand-object feature aggregation is to have the hand and object graphs fully connected (all edge), analog to \cite{doosti2020hope}. However, despite that this approach works well in sparse graphs, \emph{e.g.} including only hand joints and object bounding box corners as graph nodes, it is hard to extend the approach to a dense mesh graph. We hypothesize that the fully connected graph significantly increases the model complexity, thus making the network hard to train and converge. In addition, equally aggregating noisy features without adaptive weighting can also mislead the network in prediction.

Finally, we examine the variants where only one direction of attention is used. When we allow only hand features aggregating to object nodes (w/o object-to-hand), we observe an increase in performance in hand metrics, however, the object pose estimation is impaired compared to the converse variant (w/o hand-to-object). When the full mutual attention is included (mutual attention), we observe the best performing result. We therefore conclude that mutual attention benefits both hand and object pose estimation.

\section{Discussion}
\noindent
\textbf{Limitation.}
Our work relied on the lixel representation for hand and object meshes estimation, since the representation quantizes the image space, there is no valid correspondence to vertices that are outside the camera's field of view. Hence our method can not properly handle scenes where the hand or object is only partially included in the image. Moreover, we have only considered objects from a subset of classes where well-defined CAD models can be provided, future works should consider the interaction between hands with a more diverse set of interacting objects.

\noindent
\textbf{Conclusion. }In this paper, we proposed a novel dense mutual attention mechanism to effectively model fine-grained hand-object interaction. To exploit both intra-class and inter-class dependencies, we integrate mutual attention in the graph convolutional networks to refine the initially-estimated hand-object pose. Our method surpasses state-of-the-art 
methods when evaluated on widely-used benchmark datasets, demonstrating the effectiveness of
the proposed techniques.

{\small
\bibliographystyle{ieee_fullname}
\bibliography{egbib}
}

\end{document}